\documentclass[journal]{IEEEtran}
\usepackage{amsfonts}
\usepackage{graphicx}
\usepackage{textcomp,booktabs}
\usepackage{setspace}
\usepackage{makecell}
\usepackage{xcolor}
\graphicspath{{pic/}}
\usepackage[justification=centering]{caption}
\usepackage{amsmath}
\usepackage{url}
\usepackage{multirow}
\usepackage{threeparttable}

\ifCLASSINFOpdf
\else
\fi
\begin{document}
%
\title{DeepSeek: Paradigm Shifts and Technical Evolution in Large AI Models}
%
%
%

\author{Luolin~Xiong,~\IEEEmembership{Graduate~Student~Member,~IEEE,}
        Haofen~Wang,~\IEEEmembership{Member,~IEEE,}
        Xi~Chen,
        Lu~Sheng,~\IEEEmembership{Member,~IEEE,}
        Yun~Xiong,
        Jingping~Liu,
        Yanghua~Xiao,
        Huajun~Chen,~\IEEEmembership{Member,~IEEE,}
        Qing-Long~Han,~\IEEEmembership{Fellow,~IEEE,}
        and~Yang~Tang,~\IEEEmembership{Fellow,~IEEE}
\thanks{This work was supported by National Natural Science Foundation of China (62233005, 62293502, U2441245, 62176185, U23B2057, 62306112), the State Key Laboratory of Industrial Control Technology, China (ICT2024A22), and the Shanghai Sailing Program (23YF1409400). \emph{(Luolin Xiong, Haofen Wang, Xi Chen, Lu Sheng, Yun Xiong, Jingping Liu, Yanghua Xiao, and Huajun Chen contributed equally to this work. Corresponding author: Qing-Long Han and Yang Tang.)}}
\thanks{Luolin Xiong and Yang Tang are with the Key Laboratory of Smart Manufacturing in Energy Chemical Process, Ministry of Education, and the Engineering Research Center of Process System Engineering, Ministry of Education, East China University of Science and Technology, Shanghai 200237, China (e-mails: xiongluolin@gmail.com; tangtany@gmail.com).}
\thanks{Haofen Wang is with the College of Design and Innovation, Tongji University, Shanghai 200092, China (email: haofen.wang@tongji.edu.cn).}
\thanks{Xi Chen, Yun Xiong, and Yanghua Xiao are with the Shanghai Key Laboratory of Data Science, School of Computer Science, Fudan University, Shanghai 200433, China (emails: x\_chen21@m.fudan.edu.cn; yunx@fudan.edu.cn; shawyh@fudan.edu.cn).}
\thanks{Lu~Sheng is with the School of Software, Beihang University, Beijing 100191, China (email: lsheng@buaa.edu.cn).}
\thanks{Jingping Liu is with the School of Information Science and Engineering, East China University of Science and Technology, Shanghai 200237, China (email: jingpingliu@ecust.edu.cn).}
\thanks{Huajun Chen is with the AZFT Joint Lab for Knowledge Engine Hangzhou Innovation Center, College of Computer Science and Technology, Zhejiang University, Hangzhou 310058, China (email: huajunsir@zju.edu.cn).}
\thanks{Qing-Long~Han is with the School of Science, Computing and Engineering
Technologies, Swinburne University of Technology, Melbourne VIC 3122, Australia (email: qhan@swin.edu.au).}
}

%
%

\markboth{IEEE/CAA JOURNAL OF AUTOMATICA SINICA,~Vol.~X, No.~X, X~X}%
{Shell \MakeLowercase{\textit{et al.}}: Bare Demo of IEEEtran.cls
for Journals}
%



\maketitle

\begin{abstract}

DeepSeek, a Chinese Artificial Intelligence (AI) startup, has released their V3 and R1 series models, which attracted global attention due to their low cost, high performance, and open-source advantages. This paper begins by reviewing the evolution of large AI models focusing on paradigm shifts, the mainstream Large Language Model (LLM) paradigm, and the DeepSeek paradigm. Subsequently, the paper highlights novel algorithms introduced by DeepSeek, including Multi-head Latent Attention (MLA), Mixture-of-Experts (MoE), Multi-Token Prediction (MTP), and Group Relative Policy Optimization (GRPO). The paper then explores DeepSeek's engineering breakthroughs in LLM scaling, training, inference, and system-level optimization architecture. Moreover, the impact of DeepSeek models on the competitive AI landscape is analyzed, comparing them to mainstream LLMs across various fields. Finally, the paper reflects on the insights gained from DeepSeek’s innovations and discusses future trends in the technical and engineering development of large AI models, particularly in data, training, and reasoning.

\end{abstract}


%
\IEEEpeerreviewmaketitle

\section{introduction}
On January 20, 2025, DeepSeek released their first-generation open-weights reasoning models, DeepSeek-R1-Zero and DeepSeek-R1, which attracted worldwide attention in the AI research community and aroused significant interest from academia and industry \cite{guo2025deepseek, gibney2025china}. By January 27, the DeepSeek-R1-based chatbot application surpassed ChatGPT in downloads, becoming the most-downloaded free app on the iOS App Store in the United States \cite{WhattoKnowAboutDeepSeek:online}. This achievement contributed to a 17\% decline in Nvidia's stock price \cite{WhatisDeepSeek:online}. The DeepSeek model stands out for several reasons, including their open-source availability, reasoning performance comparable to OpenAI’s impressive o1 model, a training methodology that employs pure Reinforcement Learning (RL) without relying on Supervised Fine-Tuning (SFT) as a preliminary step, etc. Motivated by the success of DeepSeek models, it is valuable to review the technological evolution and paradigm shift of LLMs.

\begin{figure*}[t]
\centering
\includegraphics[width=0.8\linewidth]{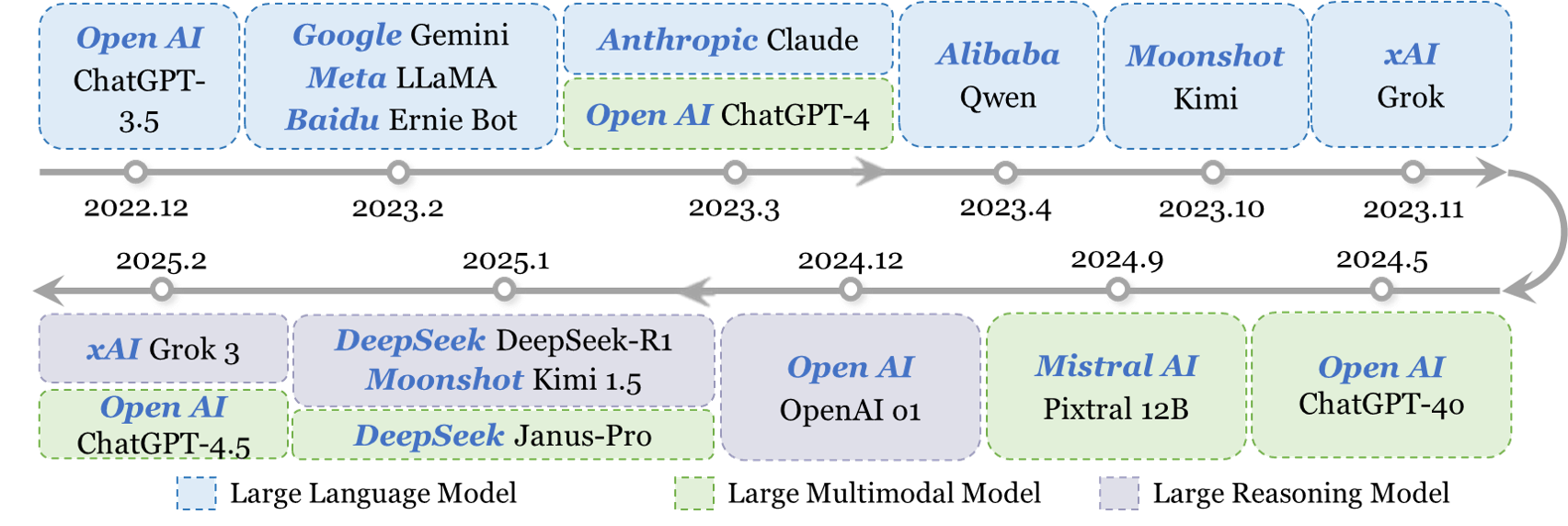}
\caption{The landscape of large AI models.}
\label{fig:company}
\end{figure*}

In recent years, a substantial amount of research and industry focus has been dedicated to large AI models, which are typically made up of deep neural networks with a large number of parameters \cite{wu2023brief, devlin2018bert, ouyang2022training, radford2021learning}. By training on massive datasets, these models are able to learn intricate representations and patterns, facilitating tasks, such as text generation \cite{li2024pretrained,min2023recent}, natural language understanding \cite{karanikolas2023large,sarikaya2014application,chowdhery2023palm}, and image recognition \cite{nadeem2024vision,bayoudh2022survey,archana2024deep}.
As advances in AI algorithms and neural network architectures continue, large AI models have demonstrated progressively enhanced data processing capabilities and exceptional performance, emerging as key drivers of industrial innovation and economic transformation.
In response, governments globally have implemented policies to foster the development of large AI models, while a multitude of technology companies have emerged to capitalize on this trend \cite{AIAction}. As shown in Fig. \ref{fig:company}, the United States leads in algorithmic innovation, computational infrastructure, and global ecosystem influence, with major players, such as OpenAI, Google, and Meta. Meanwhile, China, supported by localized needs and policy initiatives, has accelerated its efforts to close the gap and explore vertical applications, with enterprises like DeepSeek and Alibaba at the forefront. Europe, in contrast, is focused on cultivating a self-sustaining AI ecosystem.

In 2022, OpenAI launched ChatGPT, a generative AI chatbot based on LLMs, which is suitable for natural language processing tasks, such as 
question answering \cite{tan2023can,lucas2024reasoning}, conversational agents \cite{liao2023proactive,choi2024unlock,casheekar2024contemporary}, and code completion \cite{dong2024self,liu2023is}. The release of ChatGPT spurred the launch of competing products, including the Gemini \cite{team2023gemini}, Claude \cite{claude}, LLaMA \cite{meta2024introducing}, and Mistral series \cite{jiang2023mistral}. It is credited with accelerating the AI boom, which has led to ongoing rapid investment in and public attention to the field of AI. 
As the capabilities of LLMs like ChatGPT continue to evolve \cite{radford2021learning, zhuang2020comprehensive, xu2023multimodal}, researchers have begun to explore the potential of Large Multimodal Models (LMMs), which extend the scope of AI by integrating and processing multiple modalities of data, such as text, images, and audio. For example, GPT-4 has introduced the ability to process both text and image inputs \cite{achiam2023gpt}, while OpenAI's DALL-E series \cite{DALLE} can generate images from natural language descriptions. Other notable examples of multimodal models include Google DeepMind's Gemini and DeepSeek’s Janus-Pro \cite{chen2025janus}.


As task complexity continues to increase, a new frontier in LLM development emerged in late 2024, with models specifically designed for tackling challenging reasoning tasks. These advanced reasoning models spend more time generating step-by-step solutions before presenting the final answer, leading to significant enhancement in handling complex tasks, such as mathematics, scientific research, and computer programming. In 2024, Open AI released its o1 model \cite{gpto1} with reasoning capabilities and introduced inference-time scaling by extending the length of the Chain-of-Thought (CoT) reasoning process for the first time. Although it remains closed source, related technologies, such as Process Reward Models (PRM) \cite{wang2024math}, RL, and Monte Carlo Tree Search (MCTS), have been explored. On February 19, 2025, xAI introduced Grok 3 \cite{Grok3}, an advanced model blending strong reasoning with extensive pretraining knowledge. Compared with traditional LLMs, reasoning models require more computational resources.

Developing LLMs that deliver comparable performance at economical costs presents a significant challenge. In December 2024, DeepSeek, a Chinese AI company, announced the open-source release of its DeepSeek-V3 model, which has 671 billion total parameters, with 37 billion activated per token, as shown in Table \ref{tab:deepseekmodels}. The DeepSeek-V3 incorporates several innovations, including MLA for efficient inference, MoE architecture and FP8 mixed precision training for cost-effective training, and MTP for performance improving \cite{liu2024deepseek}. After thorough evaluation, DeepSeek-V3 surpasses other open-source models and delivers performance on par with leading closed-source models.


Based on DeepSeek-V3, DeepSeek-R1-Zero and DeepSeek-R1 were subsequently released. DeepSeek-R1-Zero is the first model to attempt improving reasoning ability solely through large-scale GRPO RL training, without relying on SFT and model-based rewards \cite{guo2025deepseek}. To deal with challenges, such as language mixing and poor readability in DeepSeek-R1-Zero, DeepSeek-R1 takes a two-pronged approach: it cold-starts the basic V3 model with high-quality, standardized data to improve the readability of the output, and performs SFT using both reasoning and non-reasoning synthetic data before RL to boost overall model performance \cite{guo2025deepseek}. In particular, an “Aha Moment” occurred during the training phase of DeepSeek-R1-Zero, when the model learned to allocate more time for reasoning by re-evaluating its initial approach. This behavior reflects the model’s growing reasoning abilities, and highlights that the model can autonomously develop advanced problem-solving strategies with the right incentives in RL, which contrasts to the traditional method of learning solutions directly from supervised data.

\begin{table*}[!ht]
    \centering
    \caption{Comparative analysis of DeepSeek series models.}
    \begin{tabular}{ccccc}
    \toprule
    Models & Released date  & Model parameters  & Context len. (tokens)  & Corpus size (tokens)  \\ 
    \midrule
    DeepSeek-LLM (V1) \cite{bi2024deepseek}  & Nov 2023 & 7B, 67B & 4K & 2T \\
    DeepSeek-MoE \cite{dai2024deepseekmoe}  & Jan 2024 & 16B & 4K & 2T \\
    DeepSeekMath \cite{shao2024deepseekmath} & Apr 2024 & 1.3B & 4K & 150B\\
    DeepSeek-V2 \cite{liu2024deepseekv2} & May 2024 & 236B, 21B actived per token & 128K & 8.1T \\
    DeepSeek-V3 \cite{liu2024deepseek} & Dec 2024 & 671B, 37B actived per token & 128K & 14.8T \\
    DeepSeek-R1 \cite{guo2025deepseek} & Jan 2025 & 671B & 128K & - \\
    DeepSeek-R1-Zero \cite{guo2025deepseek} & Jan 2025 & 671B & 128K & - \\
    \bottomrule
    \end{tabular}
    \label{tab:deepseekmodels}
    \begin{tablenotes}
	\footnotesize
	 \item In the table, B denotes Billion ($10^9$), K denotes Thousand ($10^3$), and T denotes Trillion ($10^{12}$).
    \end{tablenotes}
\end{table*}
The launch of DeepSeek models has not only marked about a breakthrough in AI technologies within the LLM field but also driven paradigm shift in LLM researches, with greater emphasis placed on algorithm optimization and data quality, rather than merely increasing model parameters and computational resources. This paper summarizes the paradigm shift in LLM research and analyzes the technological evolution stemming from the development of the DeepSeek series of models. The main contributions are as follows:
\begin{itemize}
    \item The mainstream LLM paradigm and its evolution are summarized. The paradigm shift introduced by the DeepSeek series models is analyzed.
\end{itemize}
\begin{itemize}
    \item The core technical innovations of DeepSeek, mainly including MLA, DeepSeekMoE, MTP, and GRPO, are presented. Additionally, engineering optimizations across the training phase, inference phase, and system-level architecture are outlined.
\end{itemize}
\begin{itemize}
    \item The impact of DeepSeek models on various fields is examined, and future development trends in LLM research are discussed.
\end{itemize}

In the remainder of this paper, the LLM paradigm and its evolution are first presented. Next, the algorithmic innovations and engineering optimizations of DeepSeek models are detailed in Section III and IV, respectively. Subsequently, the impact of DeepSeek models on various fields are analyzed, considering existing challenges and potential solutions. Lastly, the conclusion summarizes the key findings and discusses future trends in LLM research.

\section{Evolutionary History}

\begin{table*}[]
    \centering
    \caption{Paradigm Shift of LLMs}
    \footnotesize
\begin{tabular}{c|c|l|c|c|c}
\toprule
Paradigm &
  \begin{tabular}[c]{@{}c@{}}Starting \\ Time\end{tabular} &
  \multicolumn{1}{c|}{Models} &
  \begin{tabular}[c]{@{}c@{}}Model\\ Size\end{tabular} &
  \begin{tabular}[c]{@{}c@{}}Training \\ Data Scale\end{tabular} &
  Characteristic \\ \midrule
\multirow{2}{*}{\begin{tabular}[c]{@{}c@{}}Statistical and \\ Sequence Models\end{tabular}} &
  \multirow{2}{*}{2004} &
  Hidden Markov~\cite{blunsom2004hidden}, N-gram~\cite{brants2007large} &
  \begin{tabular}[c]{@{}c@{}}Up to\\ $10^{-2}$B\end{tabular} &
  \begin{tabular}[c]{@{}c@{}}Up to\\ $10^{0}$GB/$10^{-1}$B tokens\end{tabular} &
  Statistical models \\ \cmidrule{3-6} 
 &
   &
  \begin{tabular}[c]{@{}l@{}}Word2Vec~\cite{mikolov2013distributed}, GloVe~\cite{pennington2014glove}, RNN~\cite{sherstinsky2020fundamentals}, \\ LSTM~\cite{meshram2021long}\end{tabular} &
  \begin{tabular}[c]{@{}c@{}}Up to\\ $10^{-1}$B\end{tabular} &
  \begin{tabular}[c]{@{}c@{}}Up to\\ $10^1$GB/$10^{0}$B tokens\end{tabular} &
  Neural networks \\ \midrule
\multirow{2}{*}{\begin{tabular}[c]{@{}c@{}}Transformer and \\ Pre-training Models\end{tabular}} &
  \multirow{2}{*}{2018} &
  \begin{tabular}[c]{@{}l@{}}BERT~\cite{devlin2018bert}, RoBERTa~\cite{liu2019roberta}, DistilBERT~\cite{sanh2019distilbert}, \\ BioBERT~\cite{lee2020biobert}, ALBERT ~\cite{lan2019albert}, Anthropic~\cite{askell2021general}, \\ ERNIE 3.0~\cite{sun2021ernie}, T0~\cite{sanh2021multitask}\end{tabular} &
  \multirow{2}{*}{\begin{tabular}[c]{@{}c@{}}Avg. level of \\ $10^0$B,\\ Up to 65B\end{tabular}} &
  \multirow{2}{*}{\begin{tabular}[c]{@{}c@{}}Avg. level of \\ $10^1$GB/$10^0$B tokens,\\ Up to 160GB/4B tokens\end{tabular}} &
  BERT-like models \\ \cmidrule{3-3} \cmidrule{6-6} 
 &
   &
  \begin{tabular}[c]{@{}l@{}}GPT-2~\cite{radford2019language}, Codex~\cite{chen2021evaluating}, CodeGeeX~\cite{zheng2023codegeex}, \\ LLaMA~\cite{meta2023llama}, Pythia~\cite{biderman2023pythia}, Gemini 1.5~\cite{team2024gemini}\end{tabular} &
   &
   &
  GPT-like models \\ \midrule
\begin{tabular}[c]{@{}c@{}}Scaling-Up \\ of LLMs\end{tabular} &
  2019 &
  \begin{tabular}[c]{@{}l@{}}T5~\cite{2020t5}, GPT-3~\cite{brown2020language}, Gopher~\cite{rae2021scaling}, \\ BLOOM~\cite{scao2022bloom}, Chinchilla~\cite{hoffmann2022training}, GLaM~\cite{du2022glam}, \\ InstructGPT~\cite{ouyang2022training}, LaMDA~\cite{thoppilan2022lamda}, MT-NLG~\cite{smith2022using},\\ OPT~\cite{zhang2022opt}, PaLM~\cite{chowdhery2023palm}, BloombergGPT~\cite{wu2023bloomberggpt}, \\ Claude~\cite{claude}, LLaMA 2~\cite{meta2023llama}, Mistral~\cite{jiang2023mistral}, \\ LlaMa 3~\cite{dubey2024llama}, Qwen \cite{bai2023qwen}, ChatGLM \cite{glm2024chatglm}, \\ Kimi~\cite{team2025kimi}, DeepSeek-V3~\cite{liu2024deepseek}\end{tabular} &
  \multirow{3}{*}{\begin{tabular}[c]{@{}c@{}}Avg. level. of \\ $10^2$B, \\ Up to 2700B\end{tabular}} &
  \multirow{3}{*}{\begin{tabular}[c]{@{}c@{}}Avg. level of \\ $10^3$GB/$10^2$B tokens, \\ Up to 60TB/15T tokens\end{tabular}} &
  \begin{tabular}[c]{@{}c@{}}Larger models \\ (training data \\ \& parameters)\end{tabular} \\ \cmidrule{1-3} \cmidrule{6-6} 
Multimodal Learning &
  2023 &
  \begin{tabular}[c]{@{}l@{}}GPT-4~\cite{achiam2023gpt}, Gemini 2~\cite{team2023gemini}, Gemma~\cite{team2024gemma}, \\ Grok 2~\cite{grok2}, LlaMa 3.1~\cite{vavekanand2024llama}, Janus-Pro~\cite{chen2025janus}\end{tabular} &
   &
   &
  Multimodal models \\ \cmidrule{1-3} \cmidrule{6-6} 
Improved Reasoning &
  2024 &
  GPT-o1~\cite{gpto1}, DeepSeek-R1~\cite{guo2025deepseek}, Grok 3~\cite{Grok3} &
   &
   &
  Reasoning models \\ \bottomrule
\end{tabular}

\label{tab:paradigm}
\end{table*}

\subsection{LLM Paradigm Shifts}

\subsubsection{Statistical and Sequence Models}



The foundational phase of computational language modeling was dominated by statistical approaches, such as Hidden Markov Models (HMMs) \cite{blunsom2004hidden} and N-gram models \cite{brants2007large}. These methods primarily relied on counting word frequencies and calculating the probabilistic relationships between sequences of words. Although limited by their reliance on surface-level statistics, these early models provided the first insights into the probabilistic nature of language, setting the stage for more sophisticated approaches.

The next significant leap came with the advent of neural language models, which shifted the focus from discrete counting mechanisms to continuous representations of language. A landmark development in this transition was the introduction of Word2Vec \cite{mikolov2013distributed}, which pioneered the use of word embeddings. This method represented words as dense vectors in a high-dimensional space, where proximity in this space reflected semantic similarity. Building on this, GloVe \cite{pennington2014glove} introduced a model that integrated both global and local context to enhance the quality of word embeddings, further refining how words and their relationships could be modeled computationally.

With the advancement of deep learning techniques, the field saw the integration of sequential models, such as Recurrent Neural Networks (RNNs) \cite{wang2022predrnn,liu2021efficient} and Long Short-Term Memory (LSTM) networks \cite{meshram2021long}. These architectures proved invaluable for modeling sequential data, as they allowed for the retention of contextual information over long stretches of text. This capacity for handling sequential dependencies made RNNs and LSTMs crucial for advancing complex language tasks, including machine translation, speech recognition, and text generation.

\subsubsection{Emergence of Transformer and Breakthroughs in Pre-training Models}

The second stage marked a significant breakthrough with the introduction of the Transformer architecture \cite{vaswani2017attention}, which replaced earlier recurrent models with self-attention mechanisms.
Unlike RNNs and LSTMs, which process data sequentially and have difficulties handling long-range dependencies due to issues like vanishing gradients, the Transformer leverages attention mechanisms to allow direct interactions between all elements of a sequence. This ability to efficiently model long-range dependencies is one of the key advantages over traditional architectures. Moreover, the self-attention mechanism facilitates parallel processing of sequences, overcoming the sequential bottleneck of recurrent models. This not only speeds up training but also enhances the model’s ability to capture complex relationships across the entire input. These advancements led to the development of pre-trained models like Bidirectional Encoder Representations from Transformers (BERT) \cite{devlin2018bert} and Generative Pre-trained Transformer (GPT) \cite{brown2020language}, which utilized large-scale pre-training on extensive corpora to learn deep linguistic patterns before being fine-tuned for specific tasks. This era dramatically improved model performance across a range of natural language processing tasks and signaled the beginning of the large language model revolution.

The key difference between BERT and GPT lies in their training paradigms and the way they approach language understanding. BERT’s bidirectional training allows it to capture more comprehensive context around words, making it particularly effective for understanding word semantics in context. In contrast, GPT's autoregressive model excels in generating coherent and contextually relevant sequences of text, making it ideal for generative tasks. These distinct approaches reflect different philosophies in language modeling, with BERT focusing on deep understanding of context and GPT emphasizing generation of text. Both models, however, set new benchmarks in model performance across a variety of natural language processing tasks and signaled the beginning of the LLM revolution.

Both BERT-like models and GPT-like models have led to a large number of derivative models, which build on their foundational architectures. For instance, BERT-like models include RoBERTa~\cite{liu2019roberta}, DistilBERT~\cite{sanh2019distilbert}, BioBERT~\cite{lee2020biobert}, which improve on BERT’s efficiency or expand its capabilities. GPT-like models include Codex~\cite{chen2021evaluating}, LLaMA~\cite{meta2023llama}, Gemini 1.5~\cite{team2024gemini}, etc., which refining the autoregressive approach and improving generative capabilities while maintaining efficient computational requirements.


\subsubsection{Scaling-Up of LLMs}

The third stage in the development of LLMs was marked by a massive scale-up in both model size and dataset volume. Models like GPTs \cite{brown2020language, ouyang2022training}, PaLM \cite{chowdhery2023palm}, and LLaMA 2 and 3 \cite{meta2023llama, dubey2024llama}, DeepSeek-V3~\cite{liu2024deepseek}, etc., typically feature hundreds of billions or even trillions of parameters. The characteristic of these models is their ability to perform general-purpose tasks after undergoing two critical phases: initial pre-training on vast, general-purpose corpora followed by alignment with human values. Unlike previous pre-trained models, these LLMs with more training data and parameters demonstrate exceptional adaptability, seamlessly transitioning from specialized tasks to broader applications.


This stage exemplified the concept of “scaling laws”, which suggest that increasing the amount of parameters and the volume of training data results in exponential enhancements in model performance. As models grow in size, their ability to tackle a wider variety of tasks and generate more coherent, contextually aware text improves. For example, GPT-2, with 1.5 billion parameters and trained on a 40 GB huge corpus of text data, was capable of generating human-like text. However, GPT-3, with 175 billion parameters and trained on a much larger 570 GB corpus, showed a significant leap in performance, producing more fluent and contextually relevant text across a broader array of tasks, even without specialized fine-tuning.


However, the development of such super-large models comes with substantial computational requirements. Training these models requires advanced hardware such as Graphics Processing Units (GPUs) and Tensor Processing Units (TPUs), specialized for the massive parallel computation required for deep learning tasks. The training time and costs associated with these LLMs have been significant, with models like GPT-3 requiring weeks of processing on high-end hardware, leading to a growing demand for cloud-based infrastructure and distributed computing.


\subsubsection{Advancements in Model Efficiency and Multimodal Learning}

The development of LLMs has recently focused on improving efficiency without reducing performance. As training and inference costs increase with LLMs, techniques like pruning, quantization, and knowledge distillation are developed to reduce model size and computational requirements. These methods allow models to run more efficiently while retaining high-quality results. A notable example is DeepSeek-V3 \cite{liu2024deepseek}, which optimizes both performance and efficiency, making it better suited for practical, real-world applications.


Alongside these efficiency improvements, there has been a shift toward LLMs that can process various types of data—such as text, images, and audio—within the same framework. This development has enabled AI systems to understand and respond to more complex inputs. For example, GPT-4 \cite{achiam2023gpt} can handle both text and image inputs, allowing for more accurate and context-aware responses when the data types are combined. Similarly, OpenAI’s DALL-E \cite{DALLE} generates images from text descriptions, showcasing the potential of multimodal AI in creative tasks.
As multimodal models grow more complex, the need for efficient training and inference becomes even more critical. Techniques like model distillation and parameter sharing help reduce the computational cost of training while maintaining performance. This allows models to handle multimodal tasks without requiring excessive computational resources. For instance, DeepMind’s Gemini 2~\cite{team2023gemini}, Gemma~\cite{team2024gemma}, and DeepSeek’s Janus-Pro~\cite{chen2025janus} are multimodal models that balance high efficiency and strong performance. These models can handle different types of data, enabling their use in industries like healthcare, robotics, and entertainment.



\subsubsection{Improvements on Reasoning Abilities}


As LLMs are being used for more complex tasks, researchers have been focusing on improving their reasoning abilities. These tasks often require multi-step thinking, logical deduction, and decision-making over long interactions. The goal is to make these models more capable of solving problems that go beyond basic question-answering.

A major advancement in this area is GPT-o1 \cite{gpto1}, developed by OpenAI. GPT-o1 improves reasoning by expanding the CoT process. This allows the model to decompose complex problems into smaller, logical steps, mimicking how humans solve problems. Additionally, GPT-o1 introduces feedback loops. This means the model can adjust its reasoning as it works through a task, improving its focus and accuracy over time. These improvements make GPT-o1 more capable of solving tasks that need multi-step reasoning, such as proving mathematical theorems or solving puzzles. It helps the model understand context better and reduces mistakes in the reasoning process, making it more like human problem-solving.

After the success of GPT-o1, DeepSeek-R1 model \cite{guo2025deepseek} improves on this by making reasoning both faster and more efficient. While GPT-o1 made the reasoning process deeper and more accurate, R1 focuses on speeding up this process and reducing the resources needed to run the model.
R1 uses the improvements from GPT-o1 but optimizes them to work quickly and efficiently. This makes R1 a more practical solution for real-world applications, especially where computational power or time is limited. It can still solve complex problems but does so faster and with fewer resources.

\subsection{Mainstream LLM Paradigms}
The current mainstream LLMs employ advanced strategies and techniques at different stages to enhance the model’s performance, efficiency, and adaptability. Typically, LLMs consist of three key phases: pre-training, post-training, and inference and deployment. Traditional LLMs and reasoning LLMs differ in these three phases.


\subsubsection{Traditional LLMs}

In the pre-training phase, LLMs are trained on large amounts of text data to learn the basic patterns of language, such as grammar, vocabulary, and general world knowledge. Models like GPT use self-supervised learning, where they predict the next word or sentence based on the context of the text. On the other hand, BERT-like models use Masked Language Modeling (MLM), where certain words in a sentence are masked, and the model learns to guess them based on the surrounding context. Most of the LLMs use the Transformer architecture, which helps the model understand relationships between words over long distances in a sentence. The training is done using GPUs and TPUs in distributed systems to speed up the process.

Once pre-training is complete, LLMs move to the post-training phase, where they are fine-tuned on large-scale instruction data of specific tasks. This may include tasks like question answering, text classification, or dialogue generation. LLMs can also be trained to handle multiple tasks simultaneously. In certain cases, LLMs benefit from Reinforcement Learning with Human Feedback (RLHF) \cite{ouyang2022training}, which allows the models to refine their responses based on user feedback, helping to correct errors and improve performance. Additionally, LLMs are fine-tuned to reduce biases and ensure ethical responses, particularly in sensitive areas like politics, law, and medicine.



In the inference and deployment, the model generates output based on the given input, using its transformer architecture to process context and generate relevant, coherent text. Parameters like temperature can be adjusted to control output diversity. After generating the response, it may undergo post-processing to refine its fluency and formatting. In practical applications, such as chatbots or virtual assistants, LLMs continuously receive feedback from users, which helps fine-tune the model over time, enhancing its ability to adjust to new contexts and tasks.

\subsubsection{Reasoning LLMs}

In the pre-training phase, reasoning LLMs focus on enhancing reasoning capabilities and improving performance in complex, multi-step reasoning tasks. 
The pre-training process of reasoning LLMs aims to train a base model, whose pre-training process is typically similar to that of traditional LLMs, enabling the model to possess general language understanding capabilities and a certain level of world knowledge. However, to enhance its ability in complex reasoning tasks, the datasets used for reasoning LLMs may differ from those used for traditional LLMs. Reasoning LLMs are usually trained on specialized datasets that require advanced reasoning, such as scientific reasoning, legal logic, and philosophical debates. This targeted training helps the model develop stronger reasoning skills in specialized domains \cite{wang2025tutorial}.


In the post-training phase, reasoning LLMs incorporate self-feedback and error correction mechanisms. As they process tasks, they can self-validate and adjust their reasoning path to fix logical mistakes, which is a critical improvement over traditional models that lack such self-correction capabilities.
These models are trained to handle not only single-step reasoning but also multi-turn reasoning. 

To improve the model’s reasoning capabilities, the first step is to acquire reasoning step data. Traditionally, these steps may require manual annotation, but to avoid human intervention, the Self-Taught Reasoner (STaR) \cite{zelikman2022star} method is typically employed. STaR allows the model to generate intermediate reasoning steps on its own and validate these steps through its policy to providing data for subsequent training and reinforce its reasoning ability. 
For longer reasoning sequences, techniques like MCTS \cite{feng2023alphazero, luo2024improve} are used to guide the LLM to efficiently identify correct reasoning steps by exploring multiple possibilities and simulating outcomes. This is especially beneficial for tasks like solving math problems and decision-making in agent-based systems, where intermediate steps can follow different paths.

After collecting reasoning data, the next step is to train the PRM. With state transitions being deterministic and predefined, the primary objective is to develop a general reward model that directs the processes of search, reasoning, and decoding. The goal is to use a reward function to evaluate the quality of the model’s reasoning steps or final answers, 
enabling it to effectively guide the reasoning process \cite{li2022making, luo2024improve}. With the trained PRM, the LLM policy can then be optimized for enhanced reasoning. 


In the inference and deployment, unlike traditional models that perform single-step inference, reasoning LLMs generate multi-step reasoning paths dynamically. 
After training, reasoning LLMs need efficient output generation during inference. While autoregressive generation is common, more advanced techniques like beam search and MCTS are better suited for reasoning tasks \cite{wu2024empirical, snell2024scaling}. Beam search explores multiple sequences, selecting the best one based on cumulative probability, while MCTS enables the model to navigate a broader solution space by simulating and assessing various reasoning paths with a reward mechanism. The reasoning process is formalized as a Markov Decision Process (MDP) \cite{bellman1958dynamic}, where the state space represents reasoning steps, actions are potential steps or answers, the policy governs action selection, and the process-reward model guides the process by assigning rewards to actions.

In this context, CoT \cite{wei2022chain} plays a key role in structuring the reasoning process. CoT refers to the LLM’s ability to perform step-by-step reasoning, where each step is logically connected to the next \cite{wang2025tutorial}. 
CoT allows the model to autonomously perform structured reasoning, potentially exploring multiple trajectories and being guided by a PRM.
Reasoning LLMs are also designed to continuously learn from real-world feedback, and a typical RLHF method \cite{ouyang2022training} can be applied to further strengthen the model’s ability to reason effectively \cite{christianos2023pangu}.


\begin{figure*}[t]
\centering
\includegraphics[width=0.8\linewidth]{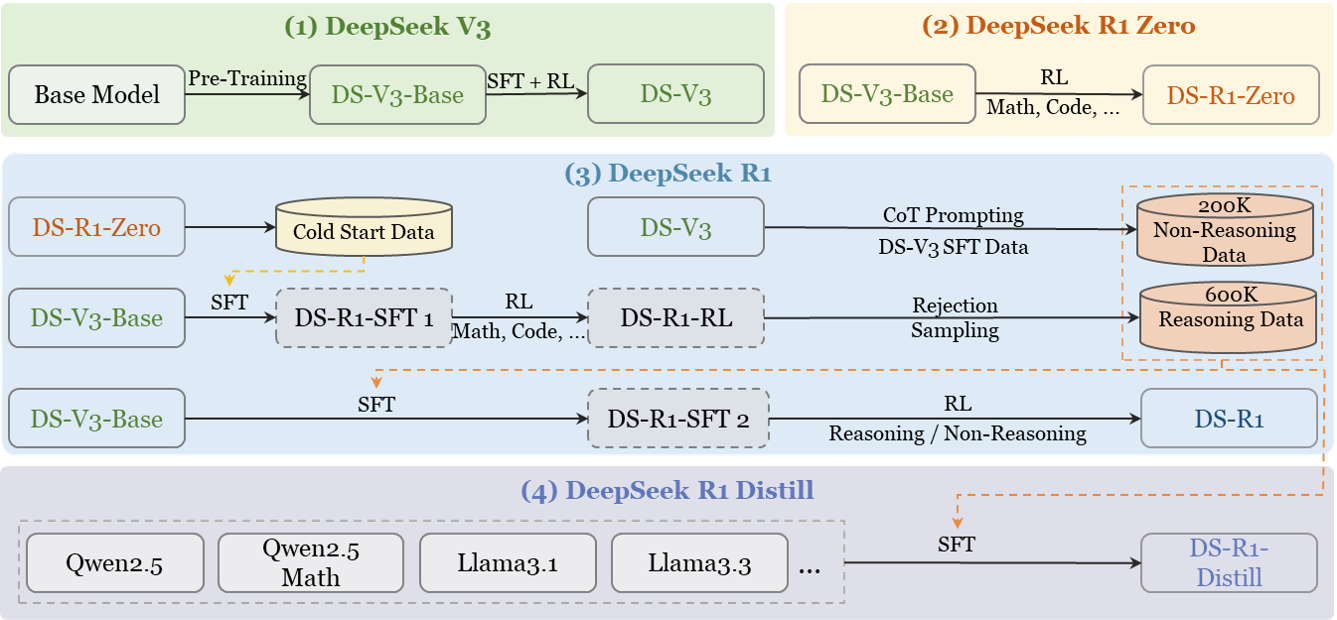}
\caption{The training pipeline of the DeepSeek (DS) series models.}
\label{fig:ds_pipeline}
\end{figure*}

\subsection{DeepSeek Paradigms}
DeepSeek has recently introduced the DeepSeek-V3~\cite{liu2024deepseek} and DeepSeek-R1 series~\cite{guo2025deepseek}. The training paradigms for these models are detailed in the official technical report and are illustrated in Fig. \ref{fig:ds_pipeline}.
Next, we provide the detail of DeepSeek V3 and DeepSeek-R1 series, respectively.

\subsubsection{DeepSeek V3}
DeepSeek-V3 is a 671B parameter, Transformer-based~\cite{vaswani2017attention} MoE model, where the training process consists of two distinct stages: pre-training and post-training.

In the pre-training phase, the base model is trained on vast corpora of 14.8 trillion diverse and high-quality tokens. These corpora incorporate several optimizations compared to the V2 version~\cite{liu2024deepseekv2}, including a higher proportion of mathematical and programming samples to enhance reasoning capabilities and expanded multilingual coverage beyond just English and Chinese.
Three key technical innovations contribute to efficient training: 1) MLA~\cite{liu2024deepseekv2} reduces the model’s Key-Value (KV) cache, which lowers memory consumption and enhances training and inference efficiency; 2) The DeepSeekMoE architecture~\cite{dai2024deepseekmoe} utilizes finer-grained experts and designates several experts as shared, helping to balance model expressiveness with computational cost control; and 3) MTP predicts multiple tokens at once, increasing the density of training signals and thereby enhancing text generation capabilities. These innovations enable the creation of the powerful DeepSeek-V3-Base model with reduced training costs.

In the post-training phase, SFT and RL are performed, resulting in DeepSeek-V3. For SFT, a 1.5M-scale multi-domain instruction dataset is synthesized using expert models (derived from internal DeepSeek-R1 variants) for reasoning tasks and DeepSeek-V2.5 for non-reasoning tasks. For RL, GRPO~\cite{shao2024deepseekmath} is employed to reduce GPU memory usage and accelerate training. Reward modeling combines rule-based and model-based approaches tailored to different problem types. This phase enhances the performance of DeepSeek-V3-Base, as well as aligning it with human preference, yielding the final DeepSeek-V3.

\subsubsection{DeepSeek-R1 series}
The DeepSeek-R1 series is a set of advanced AI models based on DeepSeek-V3-Base. The series includes three versions: DeepSeek-R1-Zero, DeepSeek-R1, and DeepSeek-R1-Distill, each designed to meet different computational and application needs.

\textbf{R1-Zero:} DeepSeek-R1-Zero is obtained by perfroming RL on V3-Base without any supervised data. 
The RL reward is modeled into two parts: accuracy reward and format reward, which respectively constrain the correctness of the answers and the format. GRPO is adopted again as the RL algorithm. For each question in each training round, several responses are sampled from the current model, and rewards are obtained respectively to get the total reward score, so as to guide the improvement of the model parameters. 
Meanwhile, this algorithm controls the update magnitude in each round and restricts the deviation from the base model (V3-Base) to improve training stability.
DeepSeek-R1-Zero has acquired a powerful reasoning ability through pure RL and performs on par with the closed-source LLM series o1 in reasoning tasks such as mathematics and code. However, there are still problems such as language mixing and poor readability, which leads to an improved training pipeline of R1.

\textbf{R1:} DeepSeek-R1 is obtained by iteratively performing SFT and RL on V3-Base, which can be divided into three stages. 

Stage 1: Collect 600K reasoning data. First, sample cold-start data with CoT from R1-Zero and perform SFT on V3-Base to form the DS-R1-SFT 1 model with basic format-following and reflection capabilities. Then, apply RL to further strengthen the reasoning ability in fields such as mathematics and code, obtaining an internal model (named DS-R1-RL). The language consistency reward is introduced to replace the accuracy reward to improve readability. Finally, 600K reasoning data are distilled from the DS-R1-RL model through rejection sampling, where samples that cannot be verified through rules are filtered using V3.

Stage 2: Collect 200K non-reasoning data generated from DeepSeek-V3 while reusing portions of its SFT dataset. For tasks that do not require reasoning, DeepSeek-V3 generates a CoT before answering, but for simpler question, no CoT is provided.

Stage 3: Train R1 through SFT and RL. Use the 800K SFT data collected in the previous two stages to perform two rounds of supervised fine-tuning on V3-Base to enhance the versatility. Then conduct reinforcement learning to obtain R1, using both reasoning and non-reasoning data, where the reasoning data, including the fields of mathematics, code and logical reasoning, uses rule-based reward, while the non-reasoning data utilizes V3-based reward model. In addition, helpfulness and harmlessness rewards are introduced to enhance the relevance of the reasoning process and reduce harmful content in the generation. Based on the complicated pipeline, DeepSeek-R1 achieves improved readability and comparable performance to advanced closed-source LLMs in both general and reasoning fields. However, the huge model parameters prevent massive users from deploying it locally, which leads to the training pipeline of distilled models.

\textbf{R1-Distill:} On a series of other open-source LLMs with relatively small parameters, SFT is performed using the 800K data collected in the R1 pipeline, significantly improving their reasoning ability and resulting in a series of DeepSeek-R1-Distill models. The base models include Qwen2.5 series~\cite{qwen2.5} and Llama-3 series~\cite{dubey2024llama} of models. Although these distilled models are inferior to R1 in performance, the computing resources needed are significantly reduced, enabling local deployment and application by individuals or development teams.

\subsubsection{DeepSeek multimodal series}

Building upon the achievements of its language models, DeepSeek has expanded its core capabilities to multimodal domains, such as visual perception and generation. In what follows, we elaborate on DeepSeek-VL2~\cite{wu2024deepseek} and Janus-Pro~\cite{chen2025janus}.

\textbf{DeepSeek-VL2:} DeepSeek-VL2 adopts a mainstream LLaVA-style~\cite{liu2024visual} framework where images are first processed by a dedicated visual encoder, such as CLIP~\cite{radford2021learning} or SigLIP~\cite{zhai2023sigmoid}, to extract rich visual features and then projected onto ``visual tokens'' through a learnable MLP.
After being concatenated with language tokens, they are jointly fed into LLMs to generate multimodal responses.
In contrast to earlier approaches, DeepSeek-VL2 introduces a dynamic tiling strategy to handle high-resolution images and extreme aspect ratios.
A high-resolution image is resized to a candidate resolution (a multiple of \(384\times384\) determined by minimal padding) and split into local \(384\times384\) tiles, with an additional global thumbnail tile for context. The pre-trained SigLIP-SO400M-384 encoder processes all tiles, generating detailed visual features with high efficiency.
Furthermore, built on the high-efficiency DeepSeekMoE architecture, DeepSeek-VL2 optimally balances visual complexity with processing speed. After training on 800B carefully curated tokens of the dataset, it achieves competitive or state-of-the-art performance on a range of multimodal benchmarks while activating a similar or even fewer number of parameters than current open-source dense and MoE-based models.

\textbf{Janus-Pro:} Janus-Pro~\cite{chen2025janus} is an enhanced version of the original Janus model~\cite{wu2024janus}, designed to excel in both multimodal understanding and text-to-image generation. Its key innovation lies in the decoupling of visual encoding for understanding and generation tasks.
Specifically, Janus-Pro employs a dedicated SigLIP encoder~\cite{zhai2023sigmoid} to extract high-dimensional semantic features for multimodal comprehension, while a separate VQ tokenizer converts images into discrete tokens for generation. These independent encoding pathways are then unified by an autoregressive transformer, allowing the model to handle diverse inputs without compromising on either task.
Complementing these methodological improvements, Janus-Pro scales both its training data and model size. Moreover, Janus-Pro has been successfully scaled to a 7B parameter variant, outperforming larger models on several benchmarks.
Collectively, these advancements position Janus-Pro as a state-of-the-art unified model, demonstrating the ability to achieve outstanding performance in both multimodal understanding and text-to-image instruction-following tasks.

\section{DeepSeek Algorithms}

In this section, we detail four key innovative algorithms in the DeepSeek series of models: MLA, DeepSeekMoE, MTP, and GRPO.

\subsection{Multi-head Latent Attention}
In DeepSeek models, the Multi-Head Attention (MHA) mechanism \cite{vaswani2017attention} in Transformer is replaced with MLA.
In the standard MHA, the input $h_t \in R^d$ is processed through three matrices $W^Q,W^K,W^V\in R^{d_{h}n_{h}\times d}$ to compute $q_t,k_t,v_t\in{R^{d_{h}n_{h}}}$. 
Instead of directly processing keys and values with large matrices, MLA \cite{liu2024deepseek} introduces compression ($W^D\in R^{d_c\times d}$) and decompression ($W^U\in R^{d_{h}n_{h}\times d_c}$) matrices. This enables the model to cache only compressed key-value pairs, reducing memory usage while preserving performance. Formally, 
\begin{equation}
\begin{aligned}
q_t=W^{Q} h_t \ \ &\to \ \  q_t=W^{UQ} (W^{DQ} h_t), \\
k_t=W^{K} h_t\ \ &\to \ \ k_t=W^{UK} (W^{DKV} h_t),    \\
v_t=W^{V} h_t\ \ &\to \ \ v_t=W^{UV} (W^{DKV} h_t),
\end{aligned}
\end{equation}
where $d_c \ll d_{h}n_{h}$. Next, Rotary Position Embedding (RoPE) \cite{su2024roformer} needs to be integrated into MLA. However, since RoPE cannot be directly applied with low-rank KV compression, DeepSeek introduces a decoupled RoPE strategy, which uses an additional query $q_t^R$ and a shared key $k_t^R$. Formally,
\begin{equation}
    \begin{aligned}
        q_t^R &={\rm \textbf{RoPE}}(W^{QR} (W^{DQ} h_t) ),   \\
        k_t^R &={\rm \textbf{RoPE}}(W^{KR} h_t ),    \\
        q_{t} &=[q_{t};q_{t}^R],  \\
        k_{t} &=[k_{t};k_t^R],
    \end{aligned}
\end{equation}
where $W^{QR}$ and $W^{KR}$ are used to generate the decoupled queries and key, respectively. ${\rm \textbf{RoPE}}(\cdot)$ represents the operation using RoPE matrices. By decoupling RoPE, MLA allows each layer to process content and position more flexibly, thereby improving model robustness to long sequences and enhancing semantic compositionality.

\textbf{Compared with MHA, MQA and GQA.} 
MHA generates $q, k, v$ by using three independent matrices \cite{2021mha,xiao2024improving}. While this approach preserves semantic information, the large KV cache limits processing size and sequence length \cite{luohe2024keep}. Multi-Query Attention (MQA) shares the same key and value across all query heads~\cite{shazeer2019fast}, reducing the KV cache by a factor of $1/h$. However, this can decrease accuracy and cause training instability \cite{brandon2024reducing}. Grouped-Query Attention (GQA)~\cite{ainslie2023gqa} further reduces the KV cache by grouping query heads, but too many groups can hurt accuracy \cite{joshi2024qcqa}. DeepSeek's MLA solves these issues by compressing the key-value pairs into latent vectors, greatly reducing the KV cache size and computational complexity without sacrificing accuracy.

\begin{figure*}[h!]
    \centering
    \includegraphics[width=0.8\textwidth]{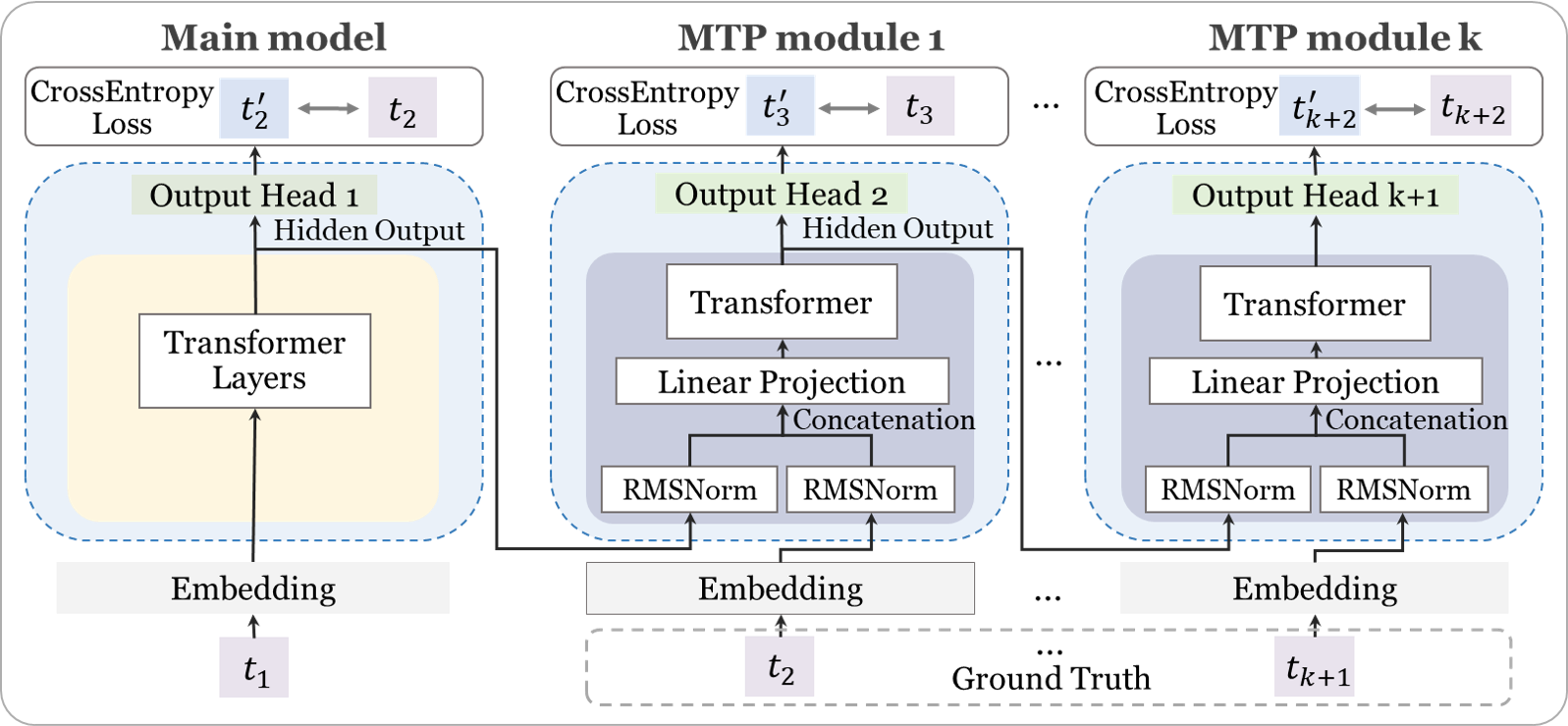}
    \caption{Illustration of Multi-Token Prediction implementation}
    \label{fig:MTP}
\end{figure*}

\subsection{DeepSeekMoE}
In DeepSeek models, the Transformer’s Feed-Forward Network (FFN) layer \cite{zhang2021moefication} is replaced by DeepSeekMoE \cite{dai2024deepseekmoe}. This architecture is built around two key optimization strategies: 1) Fine-Grained Expert Segmentation: The hidden dimensions of the FFN are partitioned into smaller, more granular experts, while preserving the total number of parameters. This enables more precise specialization among experts, enhances task-specific performance, and reduces unnecessary computation; and 2) Shared Expert Isolation: Some experts are designated as always-active shared experts, designed to capture and represent shared knowledge across various contexts. The DeepSeekMoE output $h'_t$ as follows:
\begin{equation}
    h'_t=u_t+\sum_{i=1}^{N_s}\mathrm{FFN}_i^{(s)}(u_t)+\sum_{i=1}^{N_r} g_{i,t}\mathrm{FFN}_i^{(r)}(u_t),
\end{equation}
where $u_t$ is the DeepSeekMoE input for the $t$-th token. $N_s$ and $N_r$ indicate the number of shared and routed experts, respectively. $\mathrm{FFN}_i^{(s)}\left ( \cdot  \right ) $ and $\mathrm{FFN}_i^{(r)}\left ( \cdot  \right ) $ are the $i$-th shared and routed experts, respectively. $g_{i,t}$ is the gating value for the $i$-th expert at the $t$-th token, which is defined as
\begin{equation}
g_{i,t}=\frac{g'_{i,t}}{\sum_{j=1}^{N_{r}}g'_{j,t}},
\end{equation}
\begin{equation}
g'_{i,t}=
\begin{cases}
s_{i,t}, & s_{i,t}+b_i\in\mathrm{Topk}(\{s_{j,t}+b_j|1\leq j\leq N_r\},K_r), \\
0, & \text{otherwise,}
\end{cases}
\end{equation}
where $s_{i,t}$ is the token-to-expert affinity, $\mathrm{Topk}(\cdot,K_r)$ selects the top $K$ affinity scores for the $t$-th token across routed experts, $b_i$ is a bias for each expert, used to balance load and model performance~\cite{wang2024auxiliary}.

\textbf{Compared with conventional MoE.} The existing MoE model faces two primary issues~\cite{dai2024deepseekmoe}: 1) Knowledge hybridity. Existing MoE models typically use a small number of experts (e.g., 8 and 16). This means each expert may need to handle different types of knowledge, making it hard for them to specialize in one area. As a result, the expert's parameters become too broad and less effective; and 2) Knowledge redundancy. In many cases, tokens assigned to various experts may require similar or overlapping knowledge \cite{shi2024schemoe}. As a result, multiple experts may independently learn and store the same knowledge, leading to redundancy.
DeepSeek improves upon this approach by implementing a more fine-grained expert partitioning strategy, which better captures data variations. Additionally, it introduces a dynamic routing mechanism to address the load imbalance issue commonly seen in traditional MoE models.

\subsection{Multi-Token Prediction}
Current LLMs mainly utilize a decoder-based architecture \cite{qorib2024decoder,xu2023detime}. During inference, they predict one token at a time, which causes inefficient memory access \cite{you2024linear}. DeepSeek improves on this by introducing anMTP method, where multiple future tokens are predicted at each step. Specifically, for each token $t_i$, MTP uses $D$ modules to predict $D$ additional tokens. As shown in Fig. \ref{fig:MTP} the MTP for $t_1$ is illustrated. 
 
For the $i$-th token $t_i$, the $k$-th MTP module predicts the value of its $k$-depth, denoted as $t_{i+k+1}$. It requires two inputs: the output ${h}_{i}^{k-1}$ from the $(k-1)$-th module and the embedding of the actual token $t_{i+k}$ at the previous position $i+k$. These inputs are normalized and then combined through linear projection $M_k$ to obtain $\mathbf{h}_{i}^{\prime k}$: 
\begin{equation}
\mathbf{h}_{i}^{\prime k}=M_{k}[\mathrm{RMSNorm}(\mathbf{h}_{i}^{k-1});\mathrm{RMSNorm}(\mathrm{Emb}(t_{i+k}))].
\end{equation}
Then, $\mathbf{h}_{i}^{k}$ is derived through the transformer layer $\mathrm{TRM}_k(\cdot)$. Subsequently, for the $k$-depth token prediction of $t_i$, the probability distribution of token $t_{i+k+1}$ is computed by the shared output head $\mathrm{OutHead}(\cdot)$:
\begin{equation}
\mathbf{h}_{i}^{k}=\mathrm{TRM}_{k}(\mathbf{h}_{i}^{\prime k}),
\end{equation}
\begin{equation}
P_{i+k+1}^{k}=\mathrm{OutHead}(\mathbf{h}_{i}^{k}),
\end{equation}
where $1<i+k<T$, $T$ represents the sequence length, as the prediction depth for token $t_i$ cannot exceed the sequence length.

For each MTP module, they calculate the loss $\mathcal{L}_{\mathrm{MTP}}^{k}$ using cross entropy. The average of the MTP losses across all depths is then multiplied by a weighting factor $\lambda$ to calculate the total MTP loss:
\begin{equation}
\mathcal{L}_{\mathrm{MTP}}=\frac{\lambda}{D}\sum_{k=1}^{D}\mathcal{L}_{\mathrm{MTP}}^{k}=\frac{\lambda}{D}\sum_{k=1}^{D}(-\frac{1}{T}\sum_{i=2+k}^{T}\log P_{i}^{k}[t_{i}]).
\end{equation}

\textbf{Compared with next-token prediction.} The common next-token prediction method faces training and inference bottlenecks due to inefficient memory access \cite{guo2024adaptive,yue2024object}. Gloeckle et al.~\cite{gloeckle2024better} proposed a multi-step token prediction approach that uses four parallel prediction heads connected by a shared transformer layer, allowing for the prediction of four tokens at once. However, this can cause a loss of sequential dependencies during inference \cite{bachmann2024pitfalls,tuli2024dynamo}. DeepSeek improves on this by predicting tokens sequentially, preserving the causal chain. Additionally, during training, the ground truth is included as part of the input to reduce the impact of loss.

\begin{figure*}[t!]
    \centering
    \includegraphics[width=0.8\textwidth]{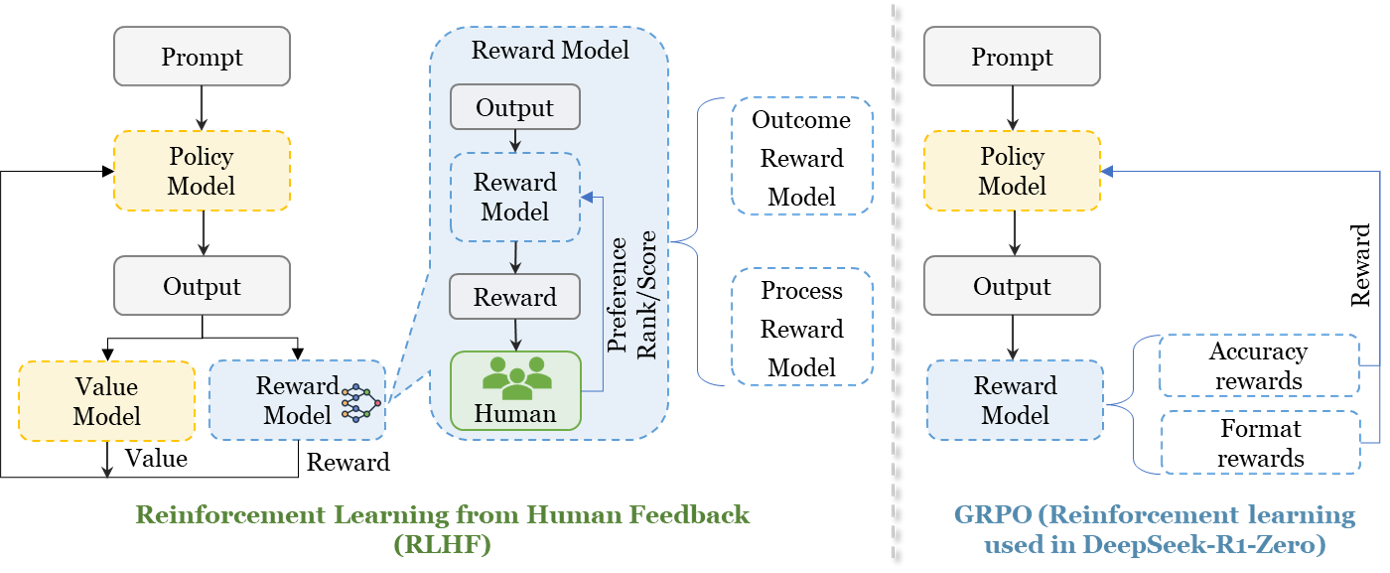}
    \caption{Comparison between RLHF and GRPO used in DeepSeek models}
    \label{fig:RL}
\end{figure*}

\subsection{Group Relative Policy Optimization}
In reinforcement learning, Proximal Policy Optimization (PPO) typically uses two models: the policy model, which generates answers, and the value model, which evaluates their quality \cite{zhang2022proximal,schulman2017proximal}. However, the value model requires a lot of resources \cite{wang2020truly} and doesn't accurately reflect the relative quality of answers due to the absolute score output \cite{gu2021proximal}. 
Therefore, DeepSeek employs the GRPO algorithm, which abandons the value model and substitutes it with accuracy reward and format reward, as depicted in Fig \ref{fig:RL}. 
Formally, for every question $q$, a set of outputs $\{ o_1,o_2, ... ,o_G \}$  is sampled from the previous policy model $\pi_{\theta_{\rm old}}$. 
These outputs are then scored using a reward model to obtain the corresponding rewards $\{r_1,r_2,…,r_G\}$. Subsequently, the rewards are normalized by subtracting the mean of the group and dividing by their standard deviation to compute the advantage $A_i$. Subsequently, the policy model output with greater advantages $o_i$ is guided by maximizing the following objective:
\begin{equation}
    A_{i}=\frac{r_{i}-mean(\{r_{1},r_{2},\cdots,r_{G}\})}{\mathrm{st}d(\{r_{1},r_{2},\cdots,r_{G}\})},
\end{equation}
\begin{equation}
\begin{aligned}
 & \mathcal{J}_{\rm GRPO}(\theta)=\mathbb{E}[q\sim P(Q),\{o_{i}\}_{i=1}^{G}\sim\pi_{\theta_{\rm old}}(O|q)] \\
 & \frac{1}{G}\sum_{i=1}^{G}\left(\min\left(I_{i}A_{i},\mathrm{clip}\left(I_{i},1-\varepsilon,1+\varepsilon\right)A_{i}\right)-\beta\mathbb{D}_{KL}\right),
\end{aligned}
\end{equation}
where $I_i=\frac{\pi_{\theta}(o_{i}|q)}{\pi_{\theta_{\rm old}}(o_{i}|q)}$. $\varepsilon$ and $\beta$ are hyper-parameters. $\pi_{\theta}(o_{i}|q)$ represents the probability that the new model will generate $o_{i}$, while $\pi_{\rm old}$ denotes the reference model.
Unlike PPO, GRPO regularizes the loss by using the Kullback-Leibler (KL) divergence directly, instead of adding KL penalties to the reward. This simplifies the computation:
\begin{equation}\mathbb{D}_{KL}\left(\pi_\theta||\pi_{\rm ref}\right)=\frac{\pi_{\rm ref}(o_{i}|q)}{\pi_{\theta}(o_{i}|q)}-\log\frac{\pi_{\rm ref}(o_{i}|q)}{\pi_{\theta}(o_{i}|q)}-1,
\end{equation}
where $\pi_{\rm ref}$ denotes the reference model.

\textbf{Compared with PPO.} PPO is a key strategy in RL, but it uses a value model that is as large as the policy model, leading to high memory and computation costs. In contrast, GRPO eliminates the value model and replaces it with a novel reward mechanism to compute group advantage scores \cite{ramesh2024group,shao2024deepseekmath}. This mechanism includes two types of rewards: 1) Accuracy rewards, which evaluate response correctness, and 2) Format rewards, which prevent content bias by enforcing structural consistency \cite{cui2025process,sane2025hybrid}. Additionally, unlike PPO, which adds KL divergence to the reward, GRPO directly incorporates it into the loss function, helping to solve the problem of ``catastrophic forgetting''.



\section{Engineering Optimization}\label{sec:eng_opt}

\subsection{The Imperative of Engineering Innovation in LLM Scaling}
The exponential scaling of LLMs precipitates fundamental engineering challenges, as compute-intensive training regimes increasingly diverge from practical deployment constraints. Empirical studies~\cite{hoffmann2022training, kaplan2020scaling} indicate that conventional scaling laws exhibit diminishing returns under current architectures, necessitating a tripartite optimization strategy encompassing algorithmic advancements, framework enhancements, and hardware co-design. By charting the developmental trajectory of DeepSeek, from foundational V3 models to reasoning-optimized R1 variants, we illustrate how co-optimization of training and inference can unlock new scaling potentials—achieving superior quality-cost efficiency for both hyperscale developers and resource-constrained environments. 

\subsection{Engineering Innovations in Training Phase}  
The training infrastructure revolution integrates precision optimization, memory orchestration, and algorithmic innovation to overcome exascale computational barriers.

DeepSeek-V3 employs a mixed precision training framework that utilizes FP8 (8-bit floating point) format for most of the model’s operations \cite{dettmers2022gpt3,noune20228,peng2023fp8}. This technique helps reduce both the computational overhead and memory usage, facilitating more efficient training without compromising model accuracy. FP8 is a lower-precision format compared to traditional BF16 or FP32, offering substantial benefits in terms of speed and memory efficiency \cite{frantar2022gptq,xiao2023smoothquant}. However, using FP8 for large-scale models usually introduces issues with outliers in the activation values, weights, and gradients, leading to potential inaccuracies \cite{fishman2024scaling, heunderstanding,sun2024massive}. To mitigate these issues, DeepSeek-V3 uses strategies like fine-grained quantization and mixed precision operations to preserve the quality of training.
Fine-grained quantization strategy ensures that the activation values are appropriately scaled, avoiding errors caused by outliers. It uses tile-wise grouping (grouping activation elements into tiles) for scaling before quantization, ensuring better handling of activation distributions.
With mixed precision operations, most computationally intensive operations (like General Matrix Multiplications, GEMM) are performed in FP8 to accelerate training, while certain critical operations (such as embedding, attention, and gating modules) remain in higher precision formats (e.g., BF16 or FP32). This balances the need for computational efficiency and numerical stability. Moreover, although FP8 is used for the majority of operations, accumulation (the process of summing intermediate results) is performed in FP32 to preserve precision and avoid underflow, which could negatively affect the training. 
The proposed FP8 mixed precision framework is validated by comparing it to BF16 training on two baseline MoE models of varying scales, with results showing that the relative error stays under 0.25\% using high-precision accumulation and fine-grained quantization techniques.
In order to achieve efficient FP8 GEMMs with fine-grained scaling, DeepSeek employs dynamic scaling adjustments, two-level accumulation to mitigate precision loss, and JIT-compiled GPU kernels optimized for NVIDIA Hopper tensor cores.
This design enables superior computational efficiency, achieving up to 2.7× speedups over optimized CUTLASS baselines while fully utilizing GPU memory bandwidth and tensor processing capabilities \cite{deepgemm2025}.

During training, DeepSeek-V3 utilizes several memory-saving techniques to reduce the overall memory footprint and enhance training efficiency. Instead of storing activations for each layer during the forward pass, the model recomputes certain intermediate values, such as RMSNorm and MLA up-projections, during backpropagation, which significantly saves memory. Additionally, the model maintains the Exponential Moving Average (EMA) of the parameters on the CPU, updating it asynchronously to avoid storing two full copies of the parameters. This approach further optimizes memory usage. Furthermore, the model shares parameters between the main model and the MTP modules, such as the embedding layer and output head, effectively reducing the memory requirements for these layers without compromising performance.


\subsection{Efficiency Breakthroughs in Inference Phase}  
The inference acceleration framework integrates attention optimization and MoE system enhancements.
One key innovation is the MLA mechanism \cite{liu2024deepseekv2}, which helps in reducing the KV cache size during inference. This is achieved by applying low-rank joint compression to the attention keys and values, which drastically reduces the memory footprint required for caching. During inference, only compressed versions of the keys and values are cached, minimizing the storage requirements. This reduction is crucial for LLMs where memory usage can become a bottleneck. Similar compression is applied to the queries in the attention mechanism, further reducing the memory footprint during inference. This is done by compressing the queries into a lower-dimensional space, optimizing memory usage without sacrificing the quality of attention calculations. 
DeepSeek also introduces a more efficient variant, FlashMLA—an optimized MLA decoding kernel for Hopper GPUs, designed for serving variable-length sequences. It achieves up to 3000 GB/s in memory-bound configurations and 580 TFLOPS in computation-bound configurations on the H800 SXM5, leveraging CUDA 12.6 \cite{flashmla2025}.
The other innovation is utilizing MoE architecture \cite{dai2024deepseekmoe, lepikhin2020gshard}. DeepSeekMoE \cite{dai2024deepseekmoe} efficiently selects a specific set of experts to handle each inference task based on the input’s needs. By activating only a few of the available experts, the model reduces the computational cost associated with processing each token. Auxiliary-loss-free load balancing ensures that expert utilization remains balanced, preventing overloading of certain experts while keeping others underused, which would otherwise reduce efficiency \cite{wang2024auxiliary, shazeer2017sparsely, fedus2022switch}.

\subsection{System-Level Optimization Architecture}
The co-designed architecture synergistically integrates distributed training infrastructure, low-level hardware optimization, and knowledge distillation paradigms to achieve unprecedented efficiency at exascale computing scales.

The DualPipe Algorithm is designed by DeepSeek to optimize Pipeline Parallelism (PP) \cite{qi2023zero} during training, a technique where the model is split into multiple stages, each running on different devices. This optimization specifically addresses the challenge of communication overhead during training, which can otherwise become a bottleneck when models scale up. 
The primary strength of the DualPipe Algorithm lies in its capacity to overlap computation and communication. In a typical PP setup, the forward pass computation and backward pass computation are done in distinct phases. Communication between stages often creates pipeline bubbles, where the next stage is waiting for data, leading to inefficiency. DualPipe solves this problem by overlapping forward and backward pass communication. The algorithm arranges the operations such that while one part of the pipeline is computing, another part is handling communication, minimizing idle time. Traditional PP can introduce delays, or pipeline bubbles, where stages of the model pipeline are idle due to waiting on data from previous stages. The DualPipe reduces these bubbles by optimizing the scheduling of data and computations, ensuring that the pipeline remains active with minimal interruptions. By optimizing how computations and communications are distributed and overlapped, DualPipe accelerates the overall training process. It minimizes the waiting time for each stage in the pipeline, increasing throughput without requiring additional computational resources. As the model scales, DualPipe helps maintain an efficient computation-to-communication ratio. This is crucial for LLMs like DeepSeek-V3, where increasing model size could otherwise lead to communication bottlenecks. The algorithm ensures that as long as the computation-to-communication ratio is maintained, scaling can continue efficiently.

Cross-Node Communication Optimization focuses on improving the communication between different nodes (computing units) in a distributed training environment, a critical challenge when training large-scale models on multiple GPUs spread across different nodes. DeepSeek-V3 optimizes this process by developing cross-node all-to-all communication kernels that leverage high-bandwidth interconnects such as InfiniBand (IB) and NVLink. These optimized communication strategies reduce the overall time spent in data transfer, allowing more time for actual computation, which speeds up the training process. NVLink provides a much higher bandwidth than traditional interconnects like InfiniBand, and DeepSeek-V3 efficiently utilizes these two technologies by allocating communication tasks to the most appropriate interconnect based on the workload. By limiting each token to a small number of nodes (no more than 4), the system reduces the communication cost while maintaining the flexibility to scale to larger numbers of experts during training. DeepSeek-V3 ensures that the load (in terms of token processing and expert assignments) is distributed efficiently across the nodes in the cluster. This load balancing helps avoid overloading certain nodes while underutilizing others, ensuring the entire system operates efficiently. Additionally, the system uses redundant experts to further enhance load balancing and improve computational efficiency by ensuring that high-load experts are effectively distributed across the nodes. The use of InfiniBand and NVLink to transfer data ensures that the communication latency is minimized. This is particularly important when dealing with large AI models and datasets, where high latency can severely slow down the training process. By optimizing how data is routed between nodes, DeepSeek-V3 ensures that communication does not become a bottleneck. With these optimizations, DeepSeek-V3 achieves near-zero overhead for all-to-all communication, allowing the model to scale effectively across multiple nodes without incurring significant communication delays. This leads to a significant reduction in training time and allows for larger-scale training without additional costs.

To further optimize communication efficiency in Mixture-of-Experts (MoE) and Expert Parallelism (EP) architectures, DeepSeek introduces DeepEP \cite{deepep2025}, a solution designed to enable high-throughput, low-latency GPU computing for large-scale distributed training and inference. DeepEP incorporates high-performance all-to-all GPU kernels for MoE dispatch and combine operations, supports low-precision FP8 computation to enhance efficiency, and integrates optimizations for DeepSeek-V3’s group-limited gating algorithm by facilitating efficient data transfer across NVLink and RDMA domains.
Additionally, DeepEP provides low-latency RDMA kernels for inference decoding and a hook-based communication overlap mechanism to further boost performance. Benchmark results indicate that FP8 precision can increase throughput by 1.5–2.5×, MoE dispatch/combine latency can be reduced by 30–50\%, and RDMA-based inference decoding can decrease end-to-end inference time by 10–25\%. These advancements make DeepEP a critical tool for large-scale distributed MoE training and inference.

For online deployment, DeepSeek employs large-scale cross-node EP to achieve higher throughput and lower latency. This method distributes experts across several GPUs, where each GPU handles a specific subset of expert tasks.
However, this approach can lead to considerable communication overhead. To address this issue, DeepSeek adopts a dual-batch overlap strategy, which reduces communication costs and boosts overall throughput by dividing a batch of requests into two microbatches.
Additionally, if a single GPU experiences excessive computational or communication load, it can become a bottleneck, slowing down the system while other GPUs remain underutilized. To optimize resource usage, DeepSeek aims to evenly distribute computational and communication workloads across all GPUs. This involves balancing core-attention computation, KVCache usage, and expert computation while ensuring an even distribution of input tokens and request counts per GPU.

\section{Influence}\label{sec:influ}

\subsection{Impact on the Large Language Model Landscape}

\begin{figure*}[t]
\centering
\includegraphics[width=0.8\linewidth]{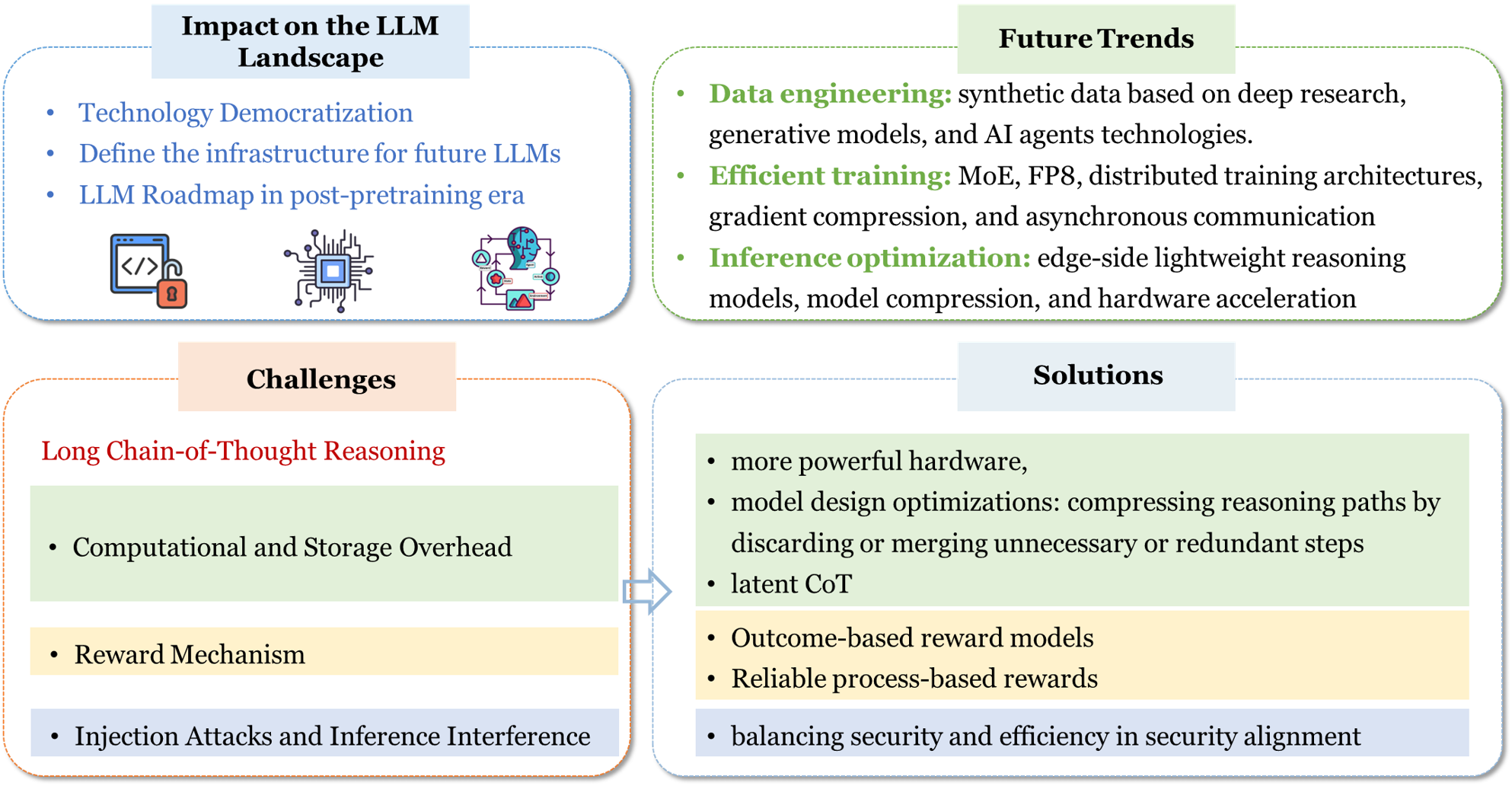}
\caption{Influence on LLM Landscape and Future Trends.}
\label{fig:future}
\end{figure*}

\noindent\textbf{Technology democratization.}
DeepSeek has fully open-sourced its model weights and inference parameters (e.g., DeepSeek-V3, R1) under the MIT license, granting global developers unrestricted rights for utilization, modification, and commercialization while maintaining copyright attribution requirements.
This move disrupts the closed-source technological monopoly, significantly reducing the barriers for small-to-medium enterprises and individual developers to access cutting-edge AI capabilities at minimal cost. It also enables LLMs to be more rapidly disseminated across various fields, helping people to obtain intelligent assistants with expert capabilities at a lower cost. Thus, it helps in technology equality and the advancement of productivity.

\vspace{+1mm}
\noindent\textbf{Define the infrastructure for future LLMs.}
DeepSeek-V3/R1 framework demonstrates the emerging potential to establish de facto industry standards through its exceptional reasoning performance and the development of open-source ecosystems. 
However, this architectural paradigm necessitates systematic reconfiguration of LLM infrastructure. 
For example, the success of the MoE architecture imposes specialized requirements for dynamic sparse computation capabilities and high-bandwidth memory. 
In addition, the significant expansion of Chain-of-Thought increases the KV cache size, resulting in a dual computational bottleneck during the intensive computation of the prefilling phase and the high-frequency memory access of the decoding phase. 
Such requirements directly catalyze the evolution of specialized chip design and software-hardware integration optimization. 
These technological advances will collectively contribute to the establishment of next-generation engineering deployment standards for LLM.

\vspace{+1mm}
\noindent\textbf{LLM roadmap in post-pretraining era.}
DeepSeek-R1 successfully utilized synthetic data and inference-time computation, significantly enhancing the LLM performance and making it possible to act as an agent to complete more complex tasks. 
Combining agents, synthetic data, and inference-time computation builds a closed loop of data generation, environment interaction, and decision optimization, jointly constructing the self-evolving AI ecosystem in the post-pretraining era. 
Integrating these three technologies creates a theoretical framework and engineering foundation for building general intelligent systems with open learning capabilities.
This paradigm can no longer rely on prepared datasets, but continuously expand the capability boundaries of LLM in dynamic and interactive environments, paving the way for the next-generation LLMs.
\subsection{Challenges}
\noindent\textbf{Long chain-of-thought reasoning.}
While DeepSeek's R1 can rival top human experts in some challenging tasks, such as code programming and solving mathematical problems, through long chain-of-thought reasoning, this kind of reasoning pattern is still in its early research phases and faces challenges:
1) Computational and Storage Overhead. As the generated reasoning trajectory increases, it will lead to a sharp increase in computational cost and energy consumption. Additionally, it poses new hardware challenges, such as efficiently managing larger key-value caches during deep reasoning.
To address this issue, two possible approaches can be considered:  
a) Beyond relying on more powerful hardware to support long reasoning processes, model design optimizations can help reduce computational overhead. This includes compressing reasoning paths by discarding or merging unnecessary or redundant steps, significantly reducing resource consumption while maintaining reasoning quality.  
b) Researchers have also explored \textit{latent CoT}~\cite{hao2024training}, which implicitly models the reasoning process, reducing computation costs. When users need to inspect the thought process, it can be decoded from the latent representation. 
2) Reward Mechanism. DeepSeek R1 currently uses outcome-based reward models (ORMs) for training, which effectively mitigate the issue of ``reward hacking'' often seen in process-based methods. However, rewarding solely based on the final answer (correct or incorrect) can overlook the correctness of intermediate reasoning steps. This poses a risk where flawed reasoning still leads to the right answer, for example, in proof-based tasks, every step must be accurate.
Therefore, how to design reliable process-based rewards while avoiding reward hacking remains an important topic for the future.

\noindent\textbf{Safety Alignment.} Models are more vulnerable to attacks in long Chain-of-Thought reasoning, where malicious manipulation can not only alter final outputs but also leak sensitive data. Studies have shown that jailbreak techniques can easily make DeepSeek generate unsafe content within reasoning chains~\cite{jiang2025safechain, ying2025reasoning}. Optimizing internal mechanisms is crucial to counter injection attacks and inference interference. However, security alignment may harm the performance. The key challenge is balancing security and efficiency: excessive alignment can degrade performance, while insufficient alignment poses risks. A flexible security strategy with dynamically adjustable parameters is needed to ensure safety while keeping performance loss within acceptable limits.

\section{conclusion and future trends}
\subsection{Summary}
As a remarkable achievement in LLM technology evolution and paradigm breakthrough, DeepSeek models introduce serval core algorithmic innovations and engineering optimizations: 1) The DeepSeek-R1-Zero model uses a pure RL, training directly on the base model without SFT. This enables the model to evolve and enhance its reasoning capabilities, particularly in complex nonlinear reasoning tasks, through self-directed exploration in the RL environment; 2) The DeepSeek series incorporates four key innovative algorithms—MLA, DeepSeekMoE, MTP, and GRPO—that enhance the model's efficiency while maintaining its performance; and 3) DeepSeek also uses mixed precision operations, where most of the heavy computations are performed in FP8 to speed up training, while important operations use higher precision formats like BF16 or FP32. These technical advantages enable it to excel in multilingual processing, logical reasoning, and computational efficiency. At the same time, DeepSeek's open-source strategy has driven the widespread application of LLMs across various fields, fostered extensive ecosystem collaboration, and progressively narrowed the gap to Artificial General Intelligence.

\subsection{future trends}
Inspired by DeepSeek's technological innovations, the future development of LLMs is expected to bring significant breakthroughs in data, training efficiency, inference optimization, and scenario-based applications. We outline several potential directions for the evolution of LLMs, as shown in Fig. \ref{fig:future}.

\subsubsection{Data engineering} In the future, LLM training will prioritize data quality, diversity, and cost, rather than focusing solely on scale. Synthetic data has emerged as a crucial and rapidly advancing tool. For instance, Deep Research efficiently extracts high-value data through semantic understanding and multimodal retrieval (text, image, video), enabling dynamic data retrieval \cite{jonesopenai}. Besides, using generative models like diffusion and GANs, synthetic data tailored to specific scenarios is automatically generated to supplement training sets. Additionally, AI agents, such as ChatGPT Operator, can develop dynamic, automated data pipelines to support the cleaning, annotation, and enhancement of streaming data, adapting to the continuous learning needs of LLMs \cite{OpenAIOperator}.

\subsubsection{Efficient training} With rapid advancements in LLM engineering, improving training efficiency—i.e., the effective use of computing resources—has become a key factor in driving further development. Techniques such as MoE help reduce computing costs and enhance algorithm efficiency. FP8 mixed precision training accelerates the training process, while distributed training architectures, combined with gradient compression and asynchronous communication strategies, minimize bandwidth consumption and improve communication efficiency. 

\subsubsection{Inference optimization} Reasoning is a critical component in LLM applications. Breakthroughs in edge-side reasoning capabilities are primarily driven by the deployment of lightweight reasoning models. Model compression techniques, like quantization, distillation, and pruning, coupled with hardware acceleration, enable the deployment of LLMs on mobile devices, Internet of Things, and other platforms. The optimization and dynamic adjustment of node relationships are particularly crucial in these lightweight models, further enhancing reasoning efficiency and accuracy by improving nonlinear expression and task-specific reasoning. For domain-specific applications, expert knowledge combined with lightweight fine-tuning can quickly tailor models to meet specific requirements.

Overall, the development of LLMs is inevitable. In the future, they may serve as the core engine driving scientific research, industrial advancement, and social progress, ultimately enabling the digital transformation of numerous industries.
\ifCLASSOPTIONcaptionsoff
  \newpage
\fi

\bibliographystyle{IEEEtran}
\bibliography{main}

\end{document}